\def\year{2021}\relax
\documentclass[letterpaper]{article} 
\usepackage{aaai21}  
\usepackage{times}  
\usepackage{helvet} 
\usepackage{courier}  
\usepackage[hyphens]{url}  
\usepackage{graphicx} 
\usepackage{amsmath}
\usepackage[switch]{lineno}  %
\usepackage{amssymb}
\usepackage{xcolor}
\usepackage{multirow}  
\usepackage{natbib}  
\usepackage{caption} 
\definecolor{localcolor}{HTML}{33A8C7}
\definecolor{syntacticcolor}{HTML}{52E3E1}
\definecolor{delimitercolor}{HTML}{A0E426}
\definecolor{nsubjcolor}{HTML}{FDF148}
\definecolor{dobjcolor}{HTML}{FFAB00}
\definecolor{amodcolor}{HTML}{F77976}
\definecolor{advmodcolor}{HTML}{F050AE}
\definecolor{blockcolor}{HTML}{D883FF}

\title{The heads hypothesis: A unifying statistical approach towards understanding multi-headed attention in BERT}

\author {
        Madhura Pande,
        Aakriti Budhraja,
        Preksha Nema,
        Pratyush Kumar, 
        Mitesh M. Khapra \\ 
}
\affiliations {
     Robert Bosch Centre for Data Science and Artificial Intelligence (RBC-DSAI) \\
    IIT Madras, India \\
   \{mpande, abudhra, preksha, pratyush, miteshk\}@cse.iitm.ac.in
}
\begin{document}
\maketitle

\begin{abstract}
    Multi-headed attention heads are a mainstay in transformer-based models. 
    Different methods have been proposed to classify the {\em role} of each attention head based on the relations between tokens which have high pair-wise attention. 
    These roles include syntactic (tokens with some syntactic relation), local (nearby tokens), block (tokens in the same sentence) and delimiter (the special \texttt{[CLS]}, \texttt{[SEP]} tokens).
    There are two main challenges with existing methods for classification: (a) there are no standard scores across studies or across functional roles, and (b) these scores are often average quantities measured across sentences without capturing statistical significance.
    In this work, we formalize a simple yet effective score that generalizes to all the roles of attention heads and employs hypothesis testing on this score for robust inference. This provides us the right lens to systematically analyze attention heads and confidently comment on many commonly posed questions on analyzing the BERT model. 
    In particular, we comment on the co-location of multiple functional roles in the same attention head, the distribution of attention heads across layers, and effect of fine-tuning for specific NLP tasks on these functional roles. Code is made publicly available. \footnote{https://github.com/iitmnlp/heads-hypothesis}
\end{abstract}

\section{Introduction}

In the short span of two years, the BERT model \cite{devlin-etal-2019-bert} has become a top contender in many NLP tasks. 
This success has led to a sub-field of research that tries to analyze why BERT works.
This sub-field has become so prominent that it has even received its own name - BERTology.
The state of BERTology is well captured by the summary in a recent survey paper: ``\textit{while BERTology has come a long way it is fair to say we still have more questions than answers about how BERT works}'' \cite{Rogers2020API}.
We see these questions as being broadly classified into three types: (a) what does BERT know, (b) what are BERT outputs sensitive to, and (c) what do the attention heads in BERT attend to?
For identifying what BERT knows, a typical strategy is to run probing experiments, for instance by evaluating POS tagging with contextual embeddings from a specific layer of BERT \cite{liu2019linguistic}.
For understanding how BERT outputs vary, ablation studies such as pruning attention heads and dropping layers are performed \cite{budhraja2020weak,michel2019sixteen,sajjad2020poor}.
Quite separate from these two questions, the third question aims to analyze the {\em intrinsic} properties of the attention heads, which are characteristic of Transformer-based models \cite{vaswani2017attention}.
Analysis and interpretation of attention patterns has been a matter of debate \cite{jain2019attention,wiegreffe2019attention}.
However, it is agreed that studying a single attention head at a time enables us to structurally localize linguistic knowledge within the model.
In this work, we survey the current approaches to analyze attention heads, identify some deficiencies, and propose a unifying and statistically robust alternative.

Multi-headed self-attention is characterised by attention weights which specify the weights of other tokens when computing the representation of the current token. 
A natural curiosity is if these attention weights encode patterns that are linguistically or structurally meaningful. 
More specifically, analysis aims to classify the role of an attention head into functional roles such as \textit{local} (attending to tokens in a small neighbourhood of the current token), \textit{syntactic} (attending to tokens which are syntactically related to the current token), \textit{delimiter} (attending to delimiter tokens such as \texttt{[CLS]} and \texttt{[SEP]}) and \textit{block} (attending to tokens within the same sentence). 
Several methods have been proposed for such classification of functional roles of attention heads. 
However, these methods vary significantly from one work to the other and also amongst the different functional roles. 
For example, \cite{kovaleva2019revealing} use the attention heatmap as an image and a CNN-based classifier to  determine whether the image exhibits a local attention pattern.
Contrast this with the method used in \cite{voita2019analyzing} which classifies a head as local if 90\% of the attention of the head lies within a small neighbourhood of the current token. 
Contrast this further with the method used in the same work to identify a syntactic role: A head is labeled as syntactic if the classification accuracy of this head on certain syntactic roles is higher than baseline classifiers.
The ``higher than'' assertion is itself based on the subjectively chosen parameter of 10\%. 

Thus, existing studies present a diverse set of methods for analysis with their respective parameters which impedes systematic comparison and evolution of ideas. 
Furthermore, existing studies have an additional challenge: statistical significance.
Clearly, as we change the input sequence (one or more sentences) to the model, the attention patterns across heads would significantly vary.
Thus, any claim on the functional role of an attention head must be shown to be consistent across the variation of input sequences.
Such analysis is well formalized in the domain of statistics where hypothesis tests must be validated with high confidence. 
However, no existing work formally presents such analysis for classification of attention heads. 
This observation is in line with the general call for more rigorous statistical validation across NLP \cite{azer2020not}.

To address these challenges with existing analysis of attention heads, we propose a single unifying metric which is then validated with hypothesis testing to determine functional roles of the attention heads.
The metric we propose is quite straightforward and is based on two observations. 
First, for every hypothesized functional role of attention heads, for a given input sequence, we can identify a target set of tokens which the current token is supposed to attend to. 
We call this target set of tokens, the attention {\em sieve}. 
Second, we need to score the preferential bias of a head in attending to tokens in a sieve.
We can aggregate the attention weights of a head to all tokens in the sieve, but we must normalize this against the number of tokens in the sieve and the total number of tokens in the input sequence. 
This normalized aggregate attention paid by an attention head to the sieve for each of the functional roles is referred to as the {\em sieve bias}.
Then, classifying an attention head to one of the functional roles can be formulated as a hypothesis test that the population mean of the corresponding sieve bias across the population of input sequences exceeds a specified threshold.
The threshold here specifies how sharp the attention bias should be to classify the functional role of a head. 
It is still subjective, but is interpretable and consistent across functional roles.

The above formulation is proposed to unify and systematize the analysis of attention heads. 
Additionally, it provides us a lens to confidently comment on many questions posed during analysis of attention heads.
We take up three main questions in this paper. 

\noindent\textbf{1. Are functional roles mutually exclusive?} 
Most existing works \cite{kovaleva2019revealing} classify attention heads to one of the different functional roles. 
This precludes the option that multiple functional roles can co-exist in the same head. 
There are two reasons to consider this possibility. 
One, functional roles defined in terms of structure (such as local heads) may overlap with syntactic roles (such as attending to \texttt {nsubj}) given that these syntactic relations are often local in the sentences.
This is a crucial overlap as recent studies have attempted to exclusively use local attention in Transformer-based models \cite{raganato2020fixed}.
Two, a single head may perform multiple roles simultaneously - e.g. attending to two different syntactic relations (such as \texttt{dobj} and \texttt{amod}).
In the paper, we quantify, visualise, and comment on such overlap in functional roles. 

\noindent\textbf{2. How are functional roles distributed across layers?} 
Existing works have shown that more syntactic attention heads are found in middle layers as opposed to other layers \cite{jawahar-etal-2019-bert,hewitt-manning-2019-structural,Goldberg2019AssessingBS}. 
Our work with hypothesis testing on sieve bias scores provides a fresh perspective on these studies. 
We can define fine-grained functional roles such as separating out the individual syntactic relations and delimiter tokens. 
Then we can quantify the fraction of heads in each layer than are {\em skilled} to perform each of these fine-grained roles. 
In this work, we quantify such skills across layers and reaffirm existing findings.

\noindent \textbf{3. What is the effect of fine-tuning on functional roles?}
A common workflow of BERT is to fine-tune the model for specific NLP tasks. 
Earlier studies \cite{kovaleva2019revealing} have shown that only the last few layers specialize during fine-tuning. 
We can analyse such specialization by quantifying the change in the {\em sieve bias} across layers and functional roles.
In this paper, we compute these changes in sieve bias across four different NLP tasks from the GLUE benchmark.

In summary, we propose a unifying approach to analyze attention heads that generalizes across functional roles and is statistically significant.
The new metrics and visualization tools enable a fresh commentary on existing findings from BERTology.

\section{Related Work}
As mentioned earlier, the field of BERTology deals with three main questions: (a) what does BERT know? \cite{lin2019open,tenney2019you,ettinger2020bert} (b) what are BERT outputs sensitive to? \cite{ethayarajh2019contextual,wiedemann2019does,mickus2019you} and (c) what do the attention heads in BERT attend to \cite{kovaleva2019revealing,clark2019does,cui2019fine}?  Of these, in this work we focus on the last question. In particular, we are interested in understanding the functional role of heads in BERT. In this section, we review existing work along two different axes.

\noindent \textbf{Types of roles identified:} Different works have identified a variety of functional roles for attention heads. These are:\\
\textit{\textbf{local:}} heads which attend to tokens in a small neighborhood around the input token, typically,  previous and/or next tokens \cite{clark2019does,kovaleva2019revealing,DBLP:journals/corr/abs-1911-12246}. Some works refer to such heads as diagonal heads \cite{kovaleva2019revealing} or positional heads \cite{voita2019analyzing}.\\
\textit{\textbf{syntactic:}} heads which attend to tokens which are syntactically related to the input token. The syntactic relations are identified using a dependency parser \cite{voita2019analyzing,clark2019does,DBLP:journals/corr/abs-1911-12246,kovaleva2019revealing,correia2019adaptively}. \\
\textit{\textbf{vertical/delimiter:}} heads which attend to the \texttt{[SEP]} and \texttt{[CLS]} tokens in the input sequence \cite{clark2019does,kovaleva2019revealing}. Some works \cite{DBLP:journals/corr/abs-1906-01698,DBLP:journals/corr/abs-1911-12246} specifically ignore the attention on \texttt{[SEP]} and \texttt{[CLS]} as these are artificially introduced tokens and not a part of the input sentence(s).\\
\textit{\textbf{block:}} heads which attend to tokens within the same sentence as opposed to tokens in other sentence before (or after) the \texttt{[SEP]} token \cite{kovaleva2019revealing}.\\
\textit{\textbf{rare word:}} heads which attend to rare words, \textit{i.e.}, words with a low frequency in the corpus \cite{voita2019analyzing}.\\
\textit{\textbf{BPE-merging:}} heads which attend to siblings of the current token resulting from the tokenisation using BPE \cite{correia2019adaptively}.\\
\textit{\textbf{interrogative:}} heads which attend to question marks at the end of a question \cite{correia2019adaptively}.\\
In this work, we focus on \textit{local}, \textit{syntactic}, \textit{block}, and \textit{delimiter} heads which are commonly studied in existing works.

\noindent \textbf{Methods used for classifying a head into a functional role:} Once the above functional roles are defined, existing works classify heads into these roles using a scoring function and an appropriate threshold. 
However, across different works there is no uniformity in the scoring function and the thresholds used for different functional roles. For example, \citeauthor{voita2019analyzing} classify a head as local if in at least 90\% of the cases, the head pays maximum attention to tokens which are immediately before or after the input token. In contrast, for determining whether a head is syntactic or not, its performance in classifying syntactic roles is compared with a simple position based baseline classifier. A head is then labelled as syntactic if its performance is at least 10\% higher than the baseline classifier.  \citeauthor{clark2019does} also follow a similar strategy wherein they classify a head as syntactic  based on its performance on identifying syntactic roles. However, unlike \citeauthor{voita2019analyzing} they do not use a threshold of 10\% but instead look for heads which substantially outperform a position based baseline classifier. \citeauthor{kovaleva2019revealing} use a CNN based classifier to assign roles to heads based on attention heatmaps. They also propose certain semantic and syntactic tests for attributing roles to heads. For example, they check if certain heads pay more attention to predicates of core frame elements as captured in FrameNet. Existing methods also differ in the treatment of special tokens such as \texttt{[SEP]} and \texttt{[CLS]}. For example, \cite{DBLP:journals/corr/abs-1911-12246} and \cite{DBLP:journals/corr/abs-1906-01698} exclude these tokens from their analysis (and thus don't consider delimiter heads). In contrast, \cite{clark2019does} show that delimiter heads are prominent in the later layers and thus important. 
Thus, the mechanisms for determining head's functional role not only varies from one work to another but also from one functional role to another. 

Another limitation with existing work is that they rely on \textit{average} functional scores of heads across a large number of input sequences. 
Since the attention patterns (e.g., of syntactic heads) vary sensitively from one input sequence to another, this raises the question of statistical significance. 
Indeed, in our experiments we find that averages are inflated by outliers sufficiently to incorrectly infer functional roles that are not statistically significant.
In this work, we address these deficiencies by proposing a statistically sound unified approach which can be adapted to any functional role.  

\section{Unifying Analysis of Attention Heads}
\subsection{The BERT model}
We analyze the BERT model which is a stack of layers each consisting of multi-headed self-attention followed by a fully connected network. 
The self-attention scheme works as follows: For every token in the input sequence an attention head transforms the token's input embedding into key, value, and query vectors, which are then linearly combined based on attention weights. 
The outputs of all heads in a layer are concatenated and then passed through the fully connected network.
The embeddings for the next layer are generated by combining the output of a layer with a skip connection from the previous layer.
We specifically consider the BERT$_{BASE}$ configuration \cite{devlin-etal-2019-bert} which has 12 layers with 12 attention heads each, and embedding vector of size 64.
\begin{figure*}
    \centering
    \includegraphics[width=7in]{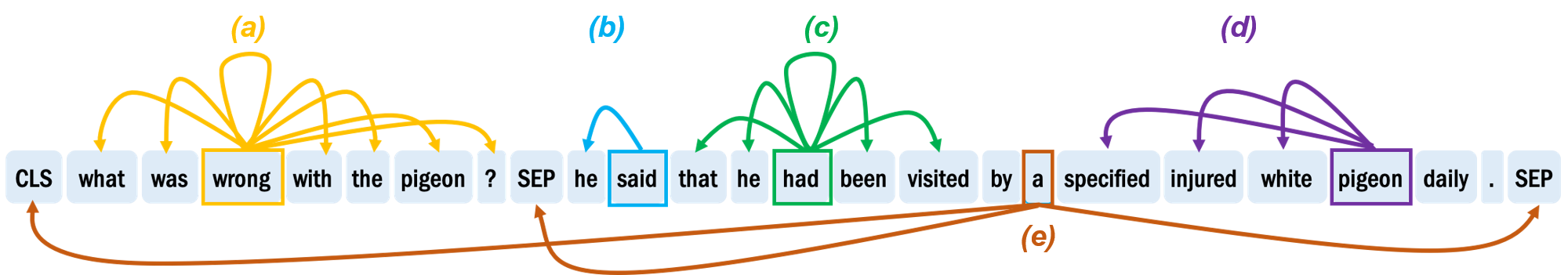}
    \caption{Attention sieves for a sample input sequence: Tokens in boxes show the current token and edges point to other tokens that are in its sieve for various functional roles: (a) block, (b) nsubj (syntactic), (c) local, (d) amod (syntactic) and, (e) delimiter.}
    \label{fig:sieves}
\end{figure*}
\subsection{Attention Sieves}
As noted, attention heads are classified to have different functional roles: local, syntactic, block, and delimiter.
Our primary aim is to define a unifying scheme to analyze attention heads across all functional roles.
The first observation is that attention heads for different functional roles are expected to attend to different sets of tokens. 
We characterize this \textit{set} of tokens for a given attention role as an attention {\em sieve}.
Formally, attention sieve $S_{I, f}(t)$ for input sequence $I$, functional role $f$, and token $t$ is the set of all tokens of $I$ that should be attended to compute the embedding of $t$ by an attention head classified to be of functional role $f$. 
We illustrate this for the example sequence shown in Figure~\ref{fig:sieves}. 
A local attention head has a clear sieve that does not depend on the content of the input sequence. 
The local attention sieve for a token are all tokens around its neighbourhood within a distance of say 2, \textit{i.e.}, 5 tokens.
For a syntactic attention head we have to first identify all tokens syntactically related to the current token with a dependency parser. 
These set of tokens would define the syntactic attention sieve.
Similarly, we define block and delimiter attention sieves.
These attention sieves can be further partitioned to consider sub-types of functional roles. 
For instance, a delimiter attention sieve can be partitioned to a \texttt{[CLS]} attention sieve and a \texttt{[SEP]} attention sieve. 
A local attention sieve can be partitioned to a prev-local or a next-local attention sieve that defines locality in only one directly.
Thus, for all tokens of an input sequence, we get attention sieves for each functional role. 

\subsection{Sieve Bias Score}
For a token $t$ of an input sequence $I$, an attention head $h$ computes an attention weight $\alpha_h(t, t')$ for each token $t' \in I$ as follows:
\begin{equation*}
    \alpha_h(t, t') = \text{softmax}\left(\dfrac{q_t^\intercal k_{t'}} {\sqrt{d_l}}\right),
    \label{eq:attn}
\end{equation*}
where $q_t$ is the query vector for token $t$, and $k_{t'}$ is the key vector for token $t'$ in the sentence. $d_l$ is the size of the embedding.
This weight captures the relative contribution of $t'$ in computing the next representation of $t$. 
If we are to posit that an attention head has a specific functional role, then attention weights to tokens in the sieve of that functional role must be higher.
Formally, if attention head $h$ is of functional role $f$ then $\sum_{t_s \in S_{I, f}(t)} \alpha_h(t, t_s)$ should be relatively large.
This aggregation of attention weights must be normalized to consider both the number of tokens in a sieve and the number of tokens in the input sequence. 
For example, the sieve of a local attention role may have 5 tokens, while that of a delimiter role may have only 2. 
Even within the same attention role, different words may be broken into varying number of tokens leading to different sizes of attention sieves.
For attention head $h$, current token $t$, input sequence $I$, and functional role $f$, we can capture this normalization with a {\em sieve bias score} as: 
\begin{equation*}
    \small
    \beta^f_h(t, I) = \left(\frac{\sum_{t_s \in S_{I, f}(t)} \alpha_h(t, t_s)}{|S_{I, f}(t)|}\right) \bigg/ \left(\frac{\sum_{t' \in I} \alpha_h(t, t')}{|I|}\right)
\end{equation*}
In the numerator, we have the average attention weight of the head to tokens within the sieve $t_s \in S_{I, f}(t)$. 
In the denominator, we have the average attention paid by the head to all tokens in the input sequence. Higher this ratio, higher is the attentive bias to the sieve.

A randomly initialised head would pay roughly equal attention weights to all tokens and thus an sieve bias score of 1 for all functional roles.
If upon learning, a head becomes more selective, then its sieve bias score would increase for specific sieves. 
Our concern then is to identify for each head the functional roles for which sieve bias scores are high. 


\begin{figure}[h]
\centering
    \includegraphics[width=2.25in]{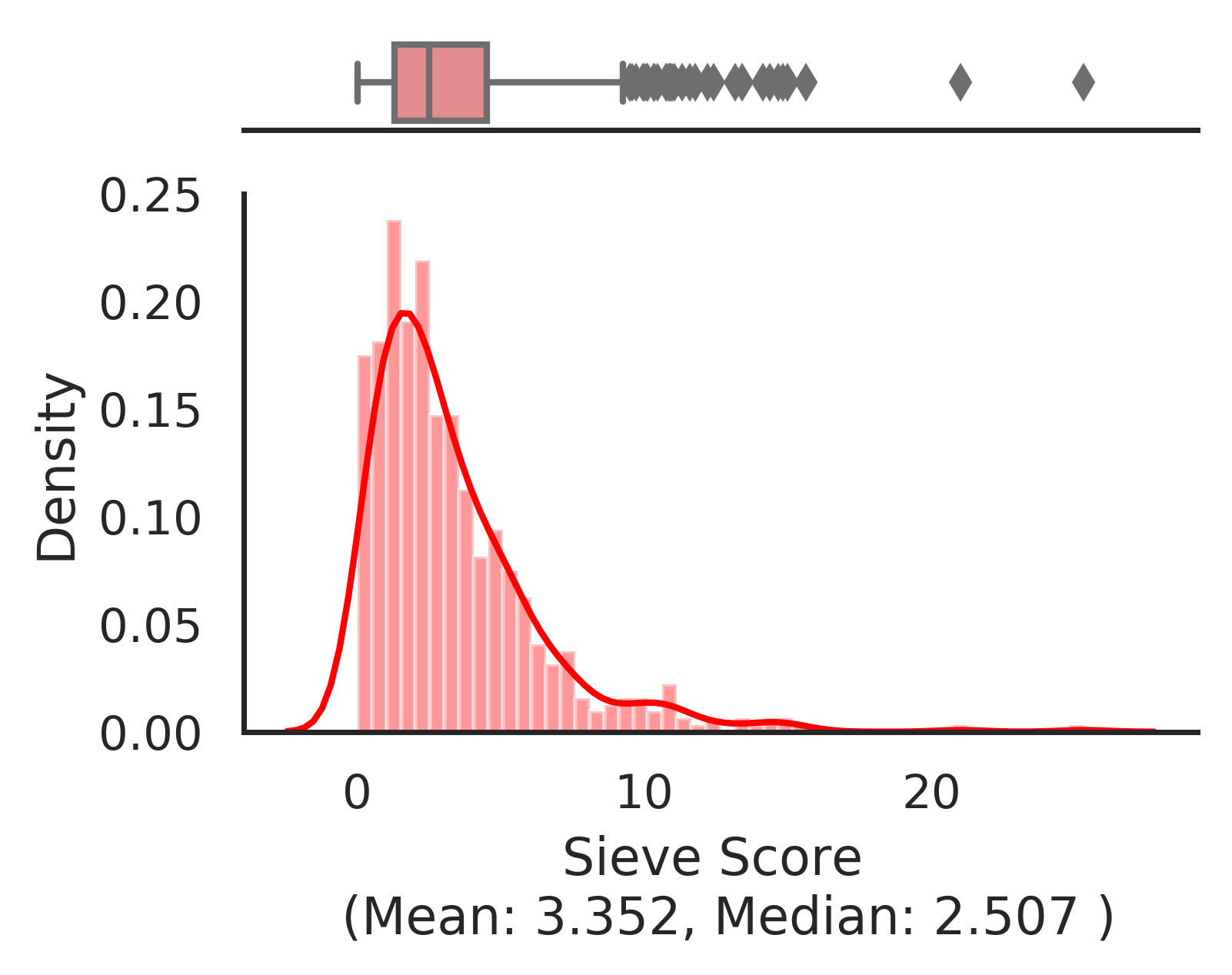}
    \caption{Sieve score distribution across sentences for a head. Note the distribution is skewed with large variation.}    
    \label{fig:token_score}
\end{figure}
\subsection{Hypothesis Testing}
We have established a relation between a head's high sieve bias score and its functional role. 
But the sieve bias scores $\beta^f_h(t, I)$ vary across input sequences. 
Can we take the average sieve bias score across many input sequences?
This would work if the sieve bias scores are symmetrically distributed with a relatively narrow standard deviation across input sequences. 
On the contrary, we found that the distributions can be skewed and broad, as shown in Figure~\ref{fig:token_score} for a specific head. 
In the presence of such variation, simple averaging is error-prone, and we instead need robust statistical inferences.
For instance, an assertion based on the average being $\geq 3$ would assign the functional role to the head in Figure~\ref{fig:token_score} . 
However, hypothesis testing to assert that population mean is more than 3 would fail.
To the best of our knowledge, no current work on analyzing attention heads considers either this variation or a statistical treatment.


We define the null hypothesis as the mean of the sieve bias score $\beta^f_h$ across all input sequences is less than or equal to some threshold $\tau$.

This threshold $\tau$ specifies how sharp the attention bias should be to classify the functional role of a head: 
Higher the value of $\tau$, stricter is the classification rule.
We can evaluate this hypothesis by using a one-tailed test of the mean with z-statistics 
, since population variance is unknown.
To this end, we can measure $\beta^f_h$ for a large number $N$ of input sequences as the sample.
We can then compute the sample mean of $\beta^f_h$ and derive the $p$-value using the one-tailed test.
As is the convention, we say that the null hypothesis is rejected if the $p$-value is smaller than 0.05.
If the null hypothesis is rejected, it follows that the attention head $h$ is satisfying the functional role $f$.
If the $p$-value is greater than $0.05$, then the test is inconclusive, and we do not ascribe the functional role to the head.

In summary, our proposed method involves the following steps: (a) computing attention sieves for all functional roles, (b) computing sieve bias scores for all pairs of attention heads and functional roles, and (c) performing hypothesis testing on the mean sieve bias score across input sequences to ascribe functional roles to each head.

\section{Results and Discussions}
In this section, we first describe our experimental setup and the datasets used for our analysis. We then provide details about the assignment of functional roles to heads using our method described earlier. Finally, based on these assigned roles, we discuss the three questions outlined earlier: i) Are  functional  roles  mutually  exclusive? ii) How are functional roles distributed across layers? and iii) What is the effect of fine-tuning on functional roles?

\subsection{Experimental Setup}
We pre-train the BERT$_{BASE}$ model using the English Wikipedia corpus and a subset of of the Project Gutenberg corpus released by \cite{lahiri:2014:SRW}.
We pre-trained the model for around 300K steps with a sequence length of $128$ and another 1K steps with a sequence length of $512$ and a batch size of 2K tokens \cite{devlin-etal-2019-bert} on a single Cloud TPU (v3-8). We then tune this model individually for four NLU tasks from the GLUE benchmark \cite{wang2018glue}: \textit{QNLI} (QA Natural Language Inference), \textit{QQP} (paraphrase detection), \textit{MRPC} (paraphrase detection) and \textit{SST-2} (sentiment analysis). For each task, we used the standard train and test splits for fine-tuning and evaluation. We used the recommended setting of hyperparameters \cite{devlin-etal-2019-bert} with batch sizes chosen among \{32, 128\} and learning rates among \{1e-4, 2e-4\}. Warm-up was set to 10K steps and LAMB optimizer was used.The performance of the fine-tuned model on these four GLUE tasks is at par with the one reported in \citeauthor{devlin-etal-2019-bert} (BERT$_{BASE}$) and suitable for further analysis. We use 1000 input sequences (num$\_$sents) from the standard test sets of the 4 GLUE tasks for all our analysis experiments. We use open-source library spaCy for dependency parsing.

\subsection{Assigning functional roles to heads}
For each task, we feed all the input sequences to the fine-tuned BERT$_{BASE}$ model and compute the sieve bias score $\beta_h^f(t, I)$ for each input sequence $I$ as described earlier. This gives us a sample of 1000 sieve bias scores (one score corresponding to each input sequence). 
Once we compute these scores for a given head $h$, and a given functional role $f$, we evaluate the null hypothesis as discussed in the earlier section based on the obtained p-value for each head. 
If the null hypothesis is rejected then it means that the head indeed performs this specific functional role. Note that we perform this hypothesis test independently for all heads and the four functional roles. This method thus allows multiple roles to be assigned to the same head. 

\begin{figure}
    \centering
    \includegraphics[width=2.25in]{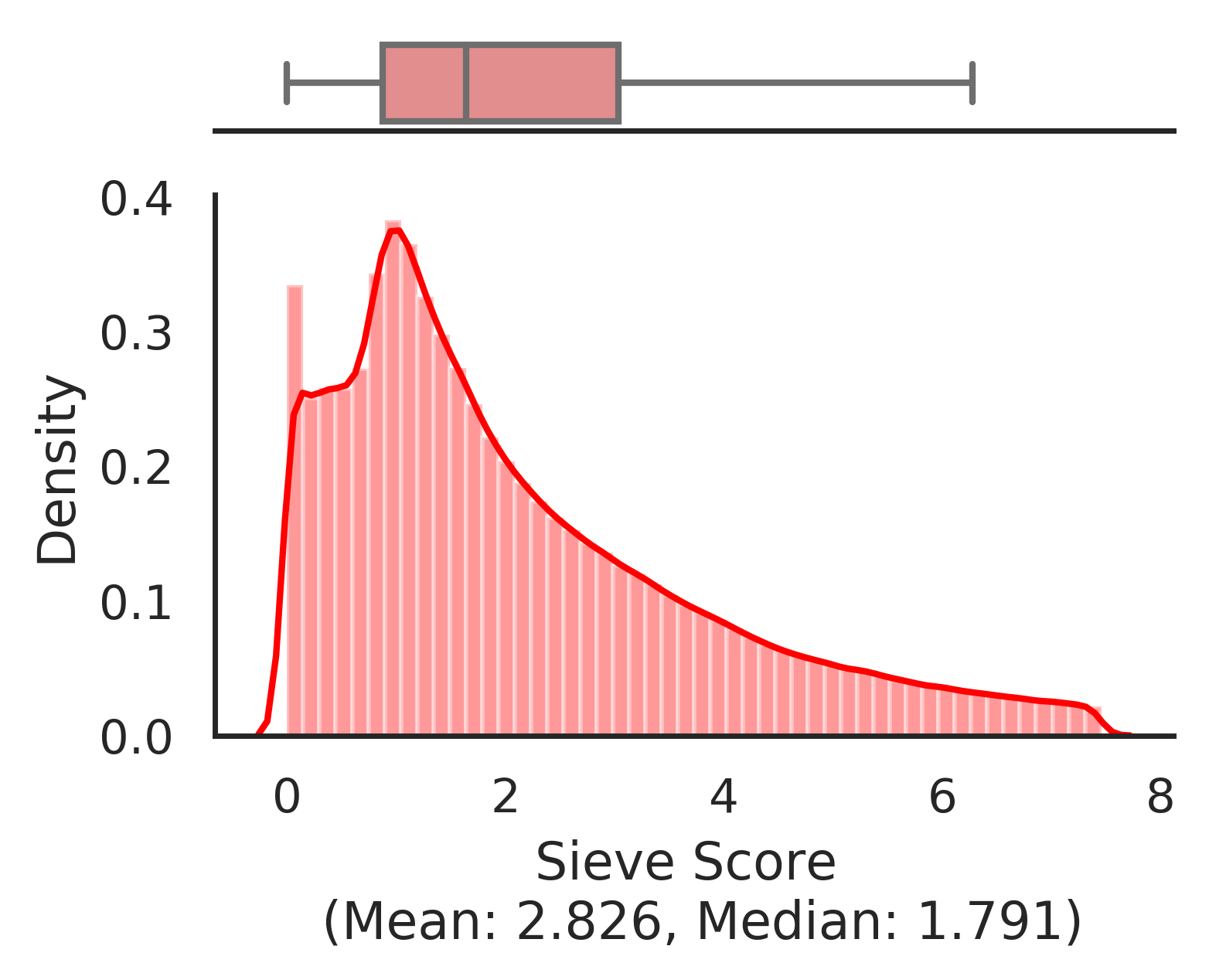} (a)
    \includegraphics[width=2.25in]{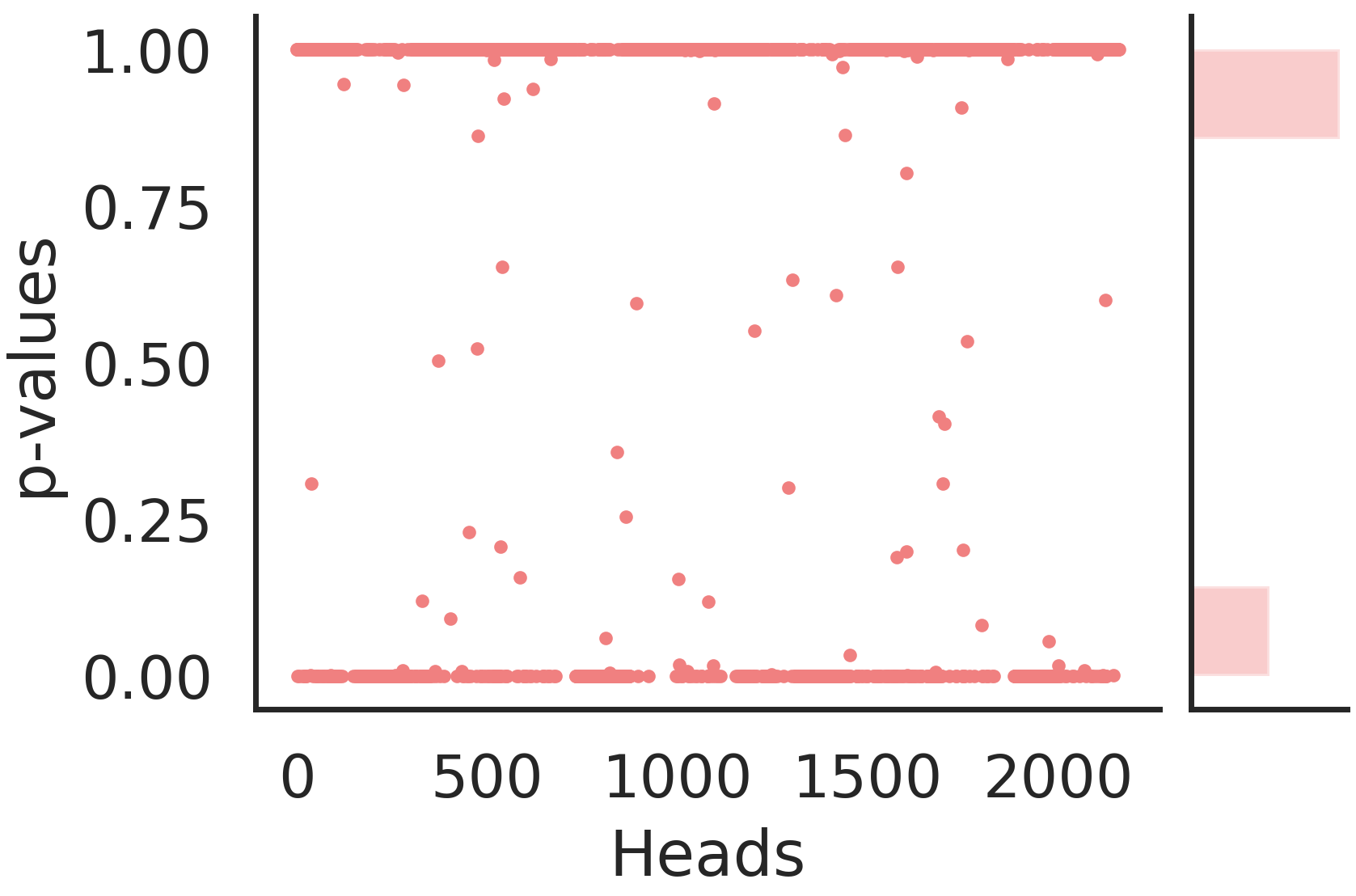} (b)
    \caption{(a) Distribution of sieve scores (144 $\times$ $|$tasks$|\times|$roles$|\times|$num\_sents$|$ scores are included) (b) p-values obtained for each head (144$\times|$tasks$|\times|$roles$|$) with hypothesis testing. High density of points on \{0,1\} indicate a high confidence level to accept/reject the Null hypothesis.}
    \label{fig:sieve_threshold_example}
\end{figure}
It is important to discuss the choice of the threshold $\tau$.
A large value of $\tau$ would lead to the null hypothesis being rejected for most heads and thus fewer functional roles assigned to heads. 
A small value of $\tau$ would lead to multiple functional roles being assigned to the same head. 
One approach to set the value of $\tau$ is to subjectively decide what constitutes \textit{attentive bias}, \textit{i.e.}, to decide the ratio of attention that a head must pay to tokens in a sieve than to all tokens.
For instance, one may suggest that $\tau = 2$ with double the attention to the sieve tokens than to other tokens is a reasonable choice for assigning functional roles to heads.
Any such value of $\tau$ can be decided and the resulting functional roles can be analysed as we do in the rest of the paper. 
In addition to this subjective check, we set the value of $\tau$ informed by data on the sieve bias scores.
Specifically, we plot the cumulative frequency distribution (see Figure~\ref{fig:sieve_threshold_example}(a)) of the sieve bias score across all functional roles for 1,000 input sequences, and identify that the mean sieve bias score is around 2.8.
We then set $\tau$ to be the smallest integer larger than this mean, i.e., 3.
We then perform the described hypothesis testing and compute the \textit{p}-values. 
We plot the individual \textit{p}-values across all the input sequences (see Figure \ref{fig:sieve_threshold_example}(b)).
We observe that this distribution is distinctly bimodal with values either close to 1 or close to 0.
Indeed, there are only 1.8\% of points in the range (0.05, 0.95). 
This indicates that the hypothesis testing made clear inferences: Either the null hypothesis was rejected (due to \textit{p}-values less than 0.05) or the test was inconclusive but with very high \textit{p}-values. 
In the former case, we assign the functional role to the head with high confidence, and in the latter case the head is clearly not satisfying that functional role as indicated by the high \textit{p}-values.
Thus, we set the threshold $\tau = 3$ to meet the two criteria: a subjective criterion of attentive bias and the objective requirement that the hypothesis testing generates clear inferences. 
Further, note that this threshold is interpretable as it simply means that the attention on the tokens within the sieve is 3 times the attention on tokens outside the sieve. 
Based on this statistically sound unified approach of classifying heads we now revisit some commonly asked questions while analysing attention heads. 

\begin{figure*}
\centering
  \includegraphics{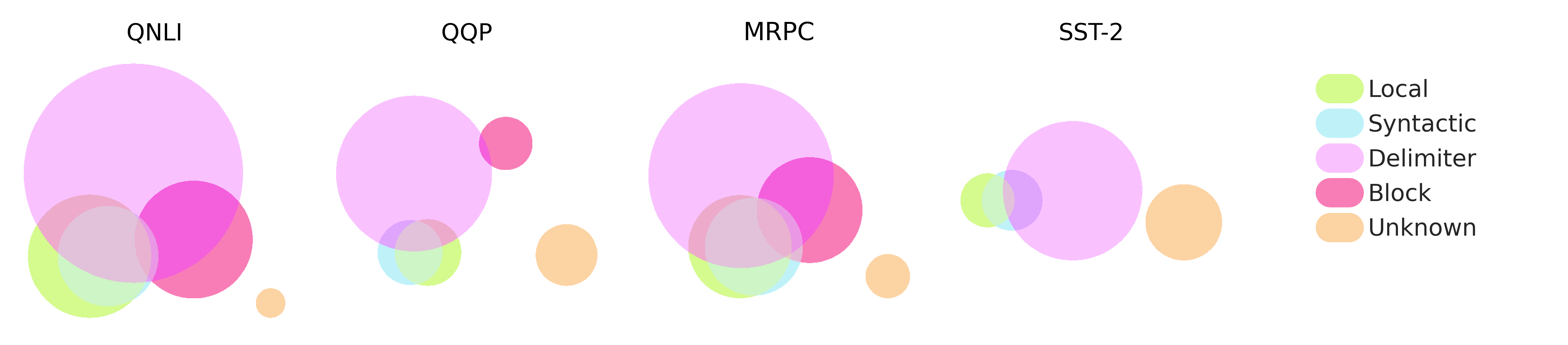}
    \caption{Venn diagram with each circle representing a set of heads with a specific role. The intersection between various sets indicate that many heads are multi-functional. Note that the size of the circle is proportional to the cardinality of the set.}
    \label{fig:heads_venn_diag}
\end{figure*}

\subsection{Are functional roles mutually exclusive?}
As mentioned earlier, existing methods typically assign only one functional role to a head. 
In contrast, as we do the hypothesis test independently for each functional role, the same head can potentially be labeled with multiple functional roles.To find out whether this is really the case, we present a Venn Diagram where heads being labeled with the same role constitute of a set (see Figure \ref{fig:heads_venn_diag}).
We make the following observations from this figure:

\noindent \textbf{Density of delimiter roles:} Across tasks, there are significant number of delimiter heads (on average 73.43\% across the 4 tasks). Also given their high prevalence, they naturally intersect with other functional heads.\\
\noindent \textbf{Unskilled heads:} There are very few heads which do not perform any functional role (avg. 9.37\% across the 4 tasks).\\
\noindent \textbf{Overlap between local and syntactic roles:} There is a significant overlap between local and syntactic heads. 
More specifically, across the four tasks, 42.1\% to 88.8\% (with a mean of 70.9\%) of syntactic heads overlap with local heads. 
This observation has an interesting connection to other studies which show that restricting attention to local patterns can still lead to good performance \cite{yang2019convolutional}. 
Our findings lend more credence to this observation, as they indicate that by attending to local tokens a head invariably also attends to syntactically related tokens. This is simply because many syntactically related tokens are in a small neighborhood around the input token. To further strengthen this claim, we compute the Spearman correlation between the local and syntactic sieve scores. We observe a very high correlation score of $0.78, 0.81, 0.85, 0.73$ scores for \textit{QNLI, QQP, MRPC, SST-2} tasks respectively, with a \textit{p}-value of $0$. We further split the syntactic and location heads based on type of syntactic relations and location of attended token (previous or next) respectively. We observe that \textit{nsubj} heads have a high overlap with \textit{next-loc} heads and \textit{dobj-heads} have a high overlap with the \textit{prev-loc} heads.\\
\noindent \textbf{Role of block heads:} There is only a marginal overlap between block heads and syntactic/local heads indicating that these specialisations are unique. Note that there are no block heads in the SST-2 task as it is a single sequence task.

 \begin{figure*}
 \centering
 \begin{tabular}{@{\hskip 0in}c}
    \includegraphics[width=7in]{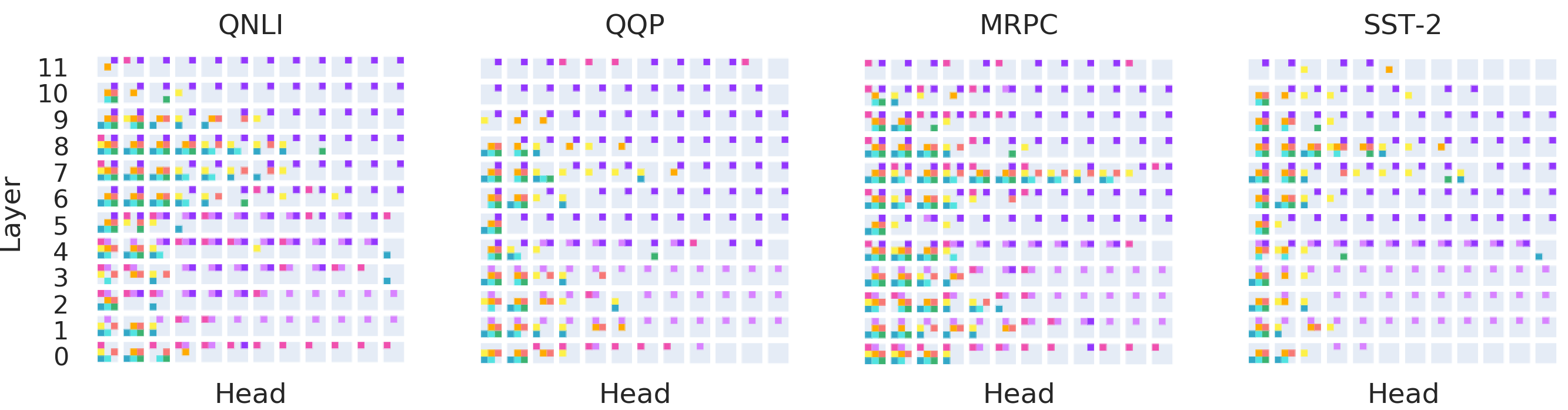} \\
    \includegraphics[width=0.8\textwidth]{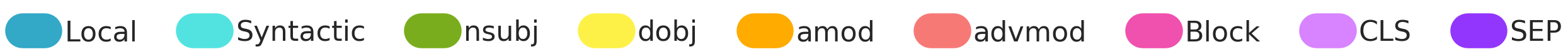}
    \end{tabular}
    \caption{Visualization of the functional roles of the attention heads of the BERT model fine-tuned on four different tasks. In each layer, the heads are arranged in a sorted order (decreasing) of their degree of multi-functionality.}
    \label{fig:labelled_heads_all}
\end{figure*}
\subsection{How are functional roles distributed across layers?}
Existing works have shown that the middle layers in BERT are more important as they contain many syntactic heads \cite{hewitt-manning-2019-structural,Goldberg2019AssessingBS,clark2019does}. Our results suggest that while this is largely true, there is significant variance across tasks. To understand this, we refer the reader to Figure \ref{fig:labelled_heads_all}.  Every larger gray colored cell in this figure corresponds to one head and each row corresponds to one layer. Thus, there are are total of 12 $\times$ 12 = 144 larger gray squares in each plot. We further divide each of these larger gray squares into 9 smaller squares.
Each of these 9 squares corresponds to a different functional role represented by a different color as defined in the legend. For this analysis, we have split the four coarse functional roles into more fine-grained functional roles. In particular, we have split the delimiter role into \texttt{[SEP]} and \texttt{[CLS]} roles. Similarly, we have split the syntactic role into \textit{nsubj, dobj, amod} and \textit{advmod} roles. Squares which are predominantly gray correspond to heads which perform very few functional roles. On the other hand, multi-colored squares correspond to \textit{multi-skilled} heads, i.e., heads performing multiple functional roles. For example, in the first plot in Figure \ref{fig:labelled_heads_all} (QNLI), 
the first head in \texttt{Layer7} performs eight different functional roles, whereas the
last head in the same layer performs only one functional role \texttt{[SEP]}. Also, to aid better visualisation, in each layer we have sorted the heads based on the number of functional roles they perform (hence, the more multi-colored (muli-functional) heads appear at the beginning of the row). Based on these plots we make the following observations: \\
\noindent\textbf{Multi-skilled heads across layers:} The number of \textit{multi-skilled} heads is higher in the middle layers (layers 5 to 9) for \textit{MRPC, QNLI} and \textit{SST-2} tasks. However, for QQP the number of multi-skilled heads is relatively higher in the initial layers. This suggests that the role of heads across layers varies from one task to another. Further, these observations may differ since earlier studies have been mostly based on analyzing a single functional role at a time.\\
\noindent\textbf{Distribution of delimiter heads:} Across tasks we find a large number of delimiter heads (\texttt{[SEP]} and/or \texttt{[CLS]}) across all layers. In fact, such delimiter heads are often the only functional heads in the last three layers across different tasks. Interestingly, the first layer contains very few (if any) delimiter heads. This may be because the delimiter tokens \texttt{[CLS]} and \texttt{[SEP]} start accumulating meaningful information from the entire sequence only after the first layer. \\
\noindent\textbf{Distribution of block heads.} For all the tasks which take two sentences as inputs (\textit{MRPC, QNLI} and \textit{QQP}) block heads are prevalent in the zeroth layer. However, their presence in the later layers varies from one task to another. For example, in both \textit{QNLI} and \textit{MRPC}, block heads consistently show up in layers 1 to 4, in addition to layer 0. However, in \textit{QQP} block heads do not show up in the middle layers but are prevalent in the first and last layer. Lastly, for \textit{MRPC} block heads consistently show up across all layers.\\
\noindent\textbf{Distribution of syntactic heads.} Across all tasks there are at least 3 syntactic heads (\textit{nsubj, dobj, amod} and \textit{advmod}) for all layers (with very few exceptions). Across tasks, the prevalence of syntactic heads is higher in layers 6 to 8 which is in line with observations made in other studies \cite{Rogers2020API, jawahar-etal-2019-bert,liu2019linguistic}.However, our findings suggest that the heads in these layers do not just perform syntactic roles but are highly \textit{multi-skilled}. Lastly, across tasks the last two layers have very few syntactic or multi-functional heads.

\subsection{What is the effect of fine-tuning on functional roles}
\begin{figure}
\begin{tabular}{@{\hskip 0in}c@{\hskip 1in}}
  \includegraphics[width=3.3in]{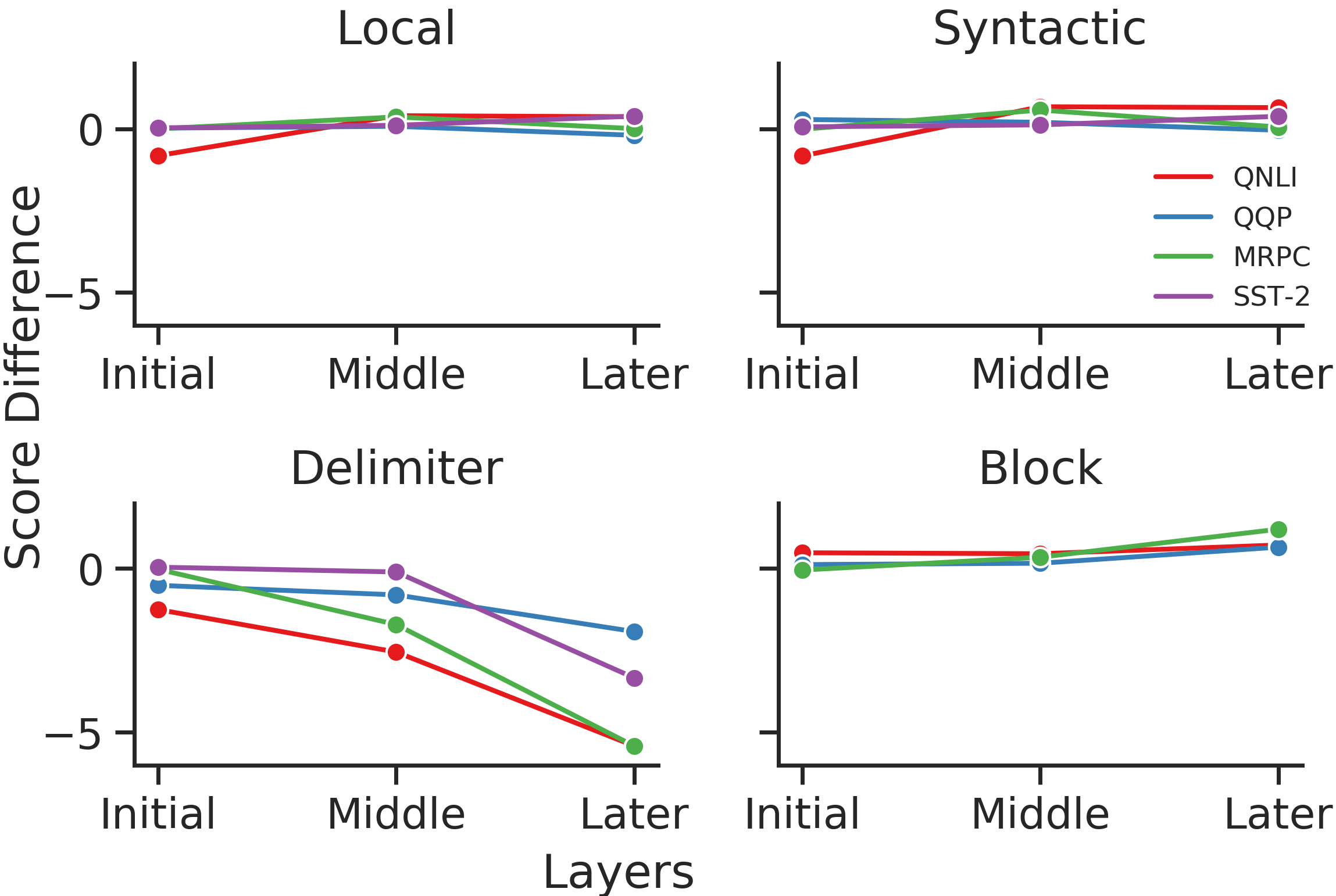}
  \end{tabular}
    \caption{Difference between the average sieve bias score after and before fine-tuning across functional roles and tasks.}
    \label{fig:mean_score_old_new}
\end{figure}
Existing works have shown that the final layers of BERT are more task specific. For example, \citeauthor{kovaleva2019revealing} show that the weights in the final layers change a lot during fine-tuning, whereas, \citeauthor{DBLP:conf/emnlp/HaoDWX19} show that the performance is less sensitive to weights in the middle layers. To substantiate these findings we measure the change in the average sieve bias scores for the 4 functional roles across initial, middle, and final sets of 4 layers for the 4 tasks (see Figure~\ref{fig:mean_score_old_new}). 
A positive (negative) value indicates that the average sieve bias score increases (decreases) after fine-tuning.
We make the following observations from these plots:

\noindent \textbf{Functional roles get redistributed in the final layers:} We observe that in the final layers, there is a decrease in the average delimiter bias score (\textit{i.e.}, avg. $\beta^{delimiter}_h$ across all heads in these layers). On further investigation, we find that this decrease is mainly due to a lower attention on the \texttt{[SEP]} but the attention on the \texttt{[CLS]} token does not change much. 
This seems to substantiate the hypothesis made in \citeauthor{clark2019does} that the \texttt{[SEP]} tokens serve as a no-op indicator and hence the attentive bias to those tokens is reducing as fine-tuning specializes the final layers.
We also find that there is an increase in the average block bias score (\textit{i.e.}, avg. $\beta^{block}_h$ across all heads in these layers). 
Thus, the later layers are specialising to generally attend to tokens within their sentence without specific syntactic bias.

\noindent \textbf{Functional roles do not change in the initial and middle layers:} In contrast to the final layers, we observe that average bias scores do not change much in the initial layers that is in line with observations made in other studies. \citeauthor{DBLP:conf/emnlp/HaoDWX19}\shortcite{DBLP:conf/emnlp/HaoDWX19} observe low sensitivity to the weights in middle layers. However, we offer a different perspective that bias scores in initial and middle layers do not change much.

\section{Conclusion}

In this work, we present a unified and formal approach for analysing the functional roles of attention heads. 
In particular, the sieve bias score generalizes across functional roles and hypothesis tests ensure statistical significance. 
This systematises the study of attention heads in BERTology.
Based on this analysis, we make several observations. 
Delimiter heads are prevalent in the network and thus overlap with other functional heads. Syntactic heads often are also local heads which explains claims made in other studies about the sufficiency of local attention patterns \cite{yang-etal-2018-modeling, yang2019convolutional, wu2019pay}. Middle layers have most number of multi-skilled heads, in line with observations made in other studies about the drop in performance while pruning heads in middle layers. Lastly, we are able to carefully study the impact of task-specific fine-tuning on the roles of the attention heads. 
We find that later layers have the largest change in their roles and specifically the number of delimiter heads drops sharply indicating specialization for task-specific roles. This again supports existing claims about the importance of later layers for specific tasks while providing a fresh perspective based on functional roles. In summary, our formal unified approach to analysing attention heads provides us the right lens to confidently comment on various questions about the functional roles of attention heads.

\section{Acknowledgements}
We thank Google for providing us with free TPUs (through TFRC program) which tirelessly ran our code day in and day out. We also thank them for supporting Preksha Nema through their Ph.D. Fellowship program. We are grateful to the Department of Computer Science, RISE Lab, Robert Bosch Center for Data Sciences and Artificial Intelligence (RBC-DSAI), and IIT Madras. We also thank the anonymous reviewers for their valuable suggestions and Madhura Pande acknowledges Maneesha for her support. 

\bibliography{main}

\end{document}